\documentclass[sigconf]{acmart}
\usepackage{url}
\usepackage{graphicx}
\usepackage{booktabs}
\usepackage{amsmath}
\usepackage{tikz}
\usepackage{rotating}

\usepackage[utf8]{inputenc}
\usepackage{pifont}
\usepackage{newunicodechar}
\newunicodechar{✓}{\ding{51}}
\newunicodechar{✗}{\ding{55}}
\usetikzlibrary{arrows,decorations.pathmorphing,backgrounds,positioning,fit,petri}
\usetikzlibrary{shadows,shapes.misc,positioning,calc}
\usetikzlibrary{decorations.pathreplacing}
\definecolor{echoreg}{HTML}{2cb1e1}
\definecolor{echodrk}{HTML}{0099cc}
\definecolor{echobg}{HTML}{eaeaea}
\definecolor{sublimedg}{HTML}{171813}
\definecolor{sublimelg}{HTML}{272822}
\definecolor{olivegreen}{rgb}{0,0.6,0}
\definecolor{myorange}{rgb}{1,0.25,0}
\definecolor{mygreen}{rgb}{0,0.6,0}
\definecolor{echodrk}{HTML}{0099cc}
\definecolor{drkorange}{HTML}{FF7c00}
\definecolor{echobg}{HTML}{eaeaea}
\definecolor{dgry}{HTML}{555555}
\definecolor{lgry}{HTML}{aaaaaa}
\definecolor{mygreen}{rgb}{0,0.6,0}
\definecolor{mygray}{rgb}{0.5,0.5,0.5}
\definecolor{mymauve}{rgb}{0.58,0,0.82}
\newcommand{\mb}{\mathbf}

\DeclareMathOperator*{\concat}{\scalebox{1}[1.5]{$\parallel$}}

\AtBeginDocument{%
  \providecommand\BibTeX{{%
    \normalfont B\kern-0.5em{\scshape i\kern-0.25em b}\kern-0.8em\TeX}}}

\definecolor{KDpurple}{rgb}{0.6,0.18,0.64}

\pdfoutput=1

\begin{document}

\title{A Bird's-Eye Tutorial of Graph Attention Architectures}

\author{Kaustubh D. Dhole}
\authornote{Corresponding Author: Suggestions and reviews are welcome.}
\email{kdhole@emory.edu}
\affiliation{%
  \institution{Emory University}
  \city{Atlanta}
  \state{Georgia}
  \country{USA}
}
\author{Carl Yang}
\email{j.carlyang@emory.edu}
\affiliation{%
  \institution{Emory University}
  \city{Atlanta}
  \state{Georgia}
  \country{USA}
}
\renewcommand{\shortauthors}{Trovato and Tobin, et al.}

\begin{abstract}
Graph Neural Networks (GNNs) have shown tremendous strides in performance for graph-structured problems especially in the domains of natural language processing, computer vision and recommender systems. Inspired by the success of the transformer architecture, there has been an ever-growing body of work on attention variants of GNNs attempting to advance the state of the art in many of these problems. Incorporating ``attention'' into graph mining has been viewed as a way to overcome the noisiness, heterogenity and complexity associated with graph-structured data as well as to encode soft-inductive bias. It is hence crucial and advantageous to study these variants from a bird's-eye view to assess their strengths and weaknesses. We provide a systematic and focused tutorial centered around attention based GNNs in a hope to benefit researchers dealing with graph-structured problems. Our tutorial looks at GNN variants from the point of view of the attention function and iteratively builds the reader's understanding of different graph attention variants.
\end{abstract}

\keywords{transformers, graph mining, graph neural networks, attention, deep learning, artificial intelligence}

\maketitle

\section{Introduction}

\begin{figure}
    \includegraphics[width=\columnwidth]{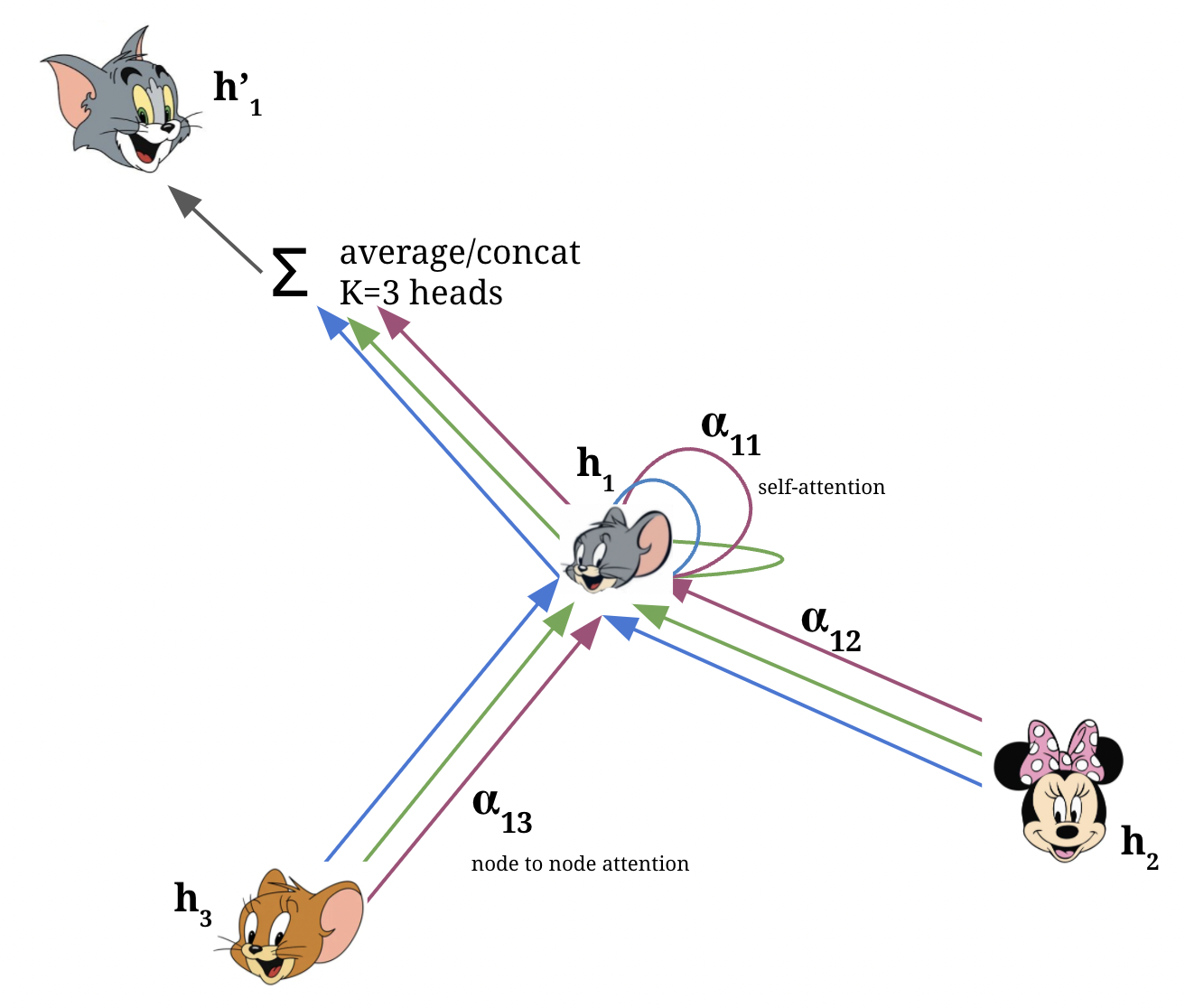}
    \caption{Multi-head attention by node 1 on its neighborhood. 3 colors depict $K=3$ heads computed individually. $\vec{h}_1'$ is obtained by concatenating or averaging features from each head.}
    \label{fig:mha}
\end{figure}

The success of deep learning, with the advent of computational power and large swathes of data has bolstered the performance of a variety of tasks, especially in the fields of natural language processing, speech recognition and computer vision. Simultaneously, there has been a proportional rise in graph-structured data in the form of knowledge-graphs, point clouds, protein and molecular data, recommender systems, etc. Unsurprisingly, there has been an increasing recent interest in extending many of the successful deep learning architectures to address the complexities associated with such ubiquitous graphical data. 

One set of such architectures referred to as graph neural networks (GNNs) have been the defacto models to cater to a plethora of problems representable in terms of nodes and edges. In Bioinformatics, GNNs have greatly benefit protein interaction prediction~\cite{dhole2014sequence} by incorporating graphical structure in addition to sequential information of proteins~\cite{yang2020graph}. GNNs have been successfully employed in the computer vision domain converting to and fro between scene graphs and images~\cite{scenegraph}, for processing point clouds~\cite{wang2019dynamic}, etc. In recommender systems, graph-based systems are popular for modelling interactions among users, products, etc.~\cite{wang2021graph} In Natural Language Processing (NLP), GNNs have been explored to interpret tree and graph representations of syntactic and semantic parses like dependency tree structures and abstract meaning representations~\cite{wang-etal-2020-amr} as well as for modelling knowledge graphs~\cite{yang2021graphformers, HittER}. Other adoptions of GNNs have spanned multifarious domains like music generation~\cite{zou2021melons}, mass spectrometry~\cite{young2021massformer}, bio-inspired camera denoising~\cite{alkendi2021neuromorphic},molecular property prediction~\cite{rong2020self}, etc. A majority of these GNNs have extended vanilla architectures of recurrent neural networks (RNNs), convolutional neural networks (CNNs), autoencoders and transformer models. 

With mammoth availability of data over the internet, real-world graphs are generally complicated, are highly heterogenous in nature and worst of all, tend to be noisy and incomplete. Attention~\cite{bahdanau2014neural} variants of GNNs have been an effective way to deal with such noise by learning to ``attend'' or ``focus'' on the relevant nodes or subgraphs while providing an empirical boost on graph tasks by encoding soft-inductive bias.

While the literature of GNNs and graph mining has been reviewed and surveyed a few times in extreme detail, most of the surveys have not significantly covered them in the context of transformers.~\cite{wu2020comprehensive} performed a comprehensive survey by proposing a 4-models taxonomy of GNNs, investigating recurrent GNNs, convolutional GNNs,
graph autoencoders, and spatial-temporal GNNs, but do not touch on attention/transformer variants.~\cite{wang2021graph} discuss graph-based representations to better recommender systems.~\cite{pitfallsofgnnevaluation} discuss some serious shortcomings of GNN evaluation.~\cite{zhou2020graph} provide a general design pipeline for GNNs and talk about architectures from a classical point of view.~\cite{cai2018comprehensive} demonstrate a systematic categorization of problems, techniques and applications of graph embeddings for more than 150 papers until 2018. The last related survey focusing on attention networks for graphs was conducted over 3 years before by ~\cite{attentioningraphs}. A plethora of attention variants have been experimented ever since viz. the GraphFormers~\cite{yang2021graphformers}, GATv2~\cite{brody2021attentive}, graph-BERT~\cite{li2022lpbert, zhang2020segmented, zhang2020graph, zhang2020g5}, LiteGT~\cite{litegt}, Graph Kernel Attention~\cite{choromanski2021graph}, Spectral Attention Network(SAN)~\cite{spectralnet} etc. It is hence crucial to survey these recent approaches to extract insights on model performance and to gauge where the field is heading.

As discussed earlier, with the rapid popularity and ground-breaking success of attention based models, many attention variants of GNNs have been experimented with by aggregating attention over other nodes of the graph~\cite{xie2020gnns} to try to further improve performance on many tasks~\cite{yang2021graphformers, brody2021attentive, li2022lpbert,zhang2020segmented, zhang2020graph, zhang2020g5, Dwivedi2021GraphNN, Yang2021GraphPN, heterogenousgraphtransfomer, choromanski2021graph}. This focused tutorial aims to discuss the different architectures of each of these later variants and their performance characteristics on downstream tasks. 

Some architectures may perform well on
certain graphs, while some many not. There is no universal architecture suitable for every problem, for the selection of the architecture
is highly dependent on the traits of the graphs. Our aim is to equip readers with a thorough understanding of these architectures to help contextualise them in their particular problem. While this review is primarily targeted towards graph practitioners, we are optimistic that newbies who are curious about graph neural networks will be able to make the best use of the same.

In Section 2, we first introduce a high-level taxonomy of graph problems. In Section 3, we discuss the limitations of the message-passing paradigm. In Section 4, we introduce the attention architectures used and lead the reader gradually through each of the attention variants starting from the Veličković et al~\cite{velivckovic2017graph}’s seminal work.

\section{Taxonomy of Graph Problems}
We taxonomize attention variants of GNNs according to four intended downstream applications viz. node level problems, edge level problems, graph level problems and others which do not fall under the former three.
The tasks in the bracket describe the datasets/tasks over which each of the architectures has been evaluated by the original authors or benchmarked in subsequent papers.

\subsection{Node Level}
Node level tasks majorly include supervised tasks like node classification and node regression as well as unsupervised tasks like node clustering. Knowledge graph problems involving prediction of labels of new nodes would be a popular application. Model performance has been gauged using a plethora of benchmark datasets namely Cora, Citeseer, Pubmed, etc. In this survey, the following node classification architectures would be discussed. 
\begin{itemize}
    \item GAT (Cora, Citeseer, PubMed, PPI)
    \item GATv2 (Cora, Citeseer, PubMed, PPI)
    \item GaAN (PPI~\cite{hamilton2017inductive}, Reddit~\cite{hamilton2017inductive}, METR-LA~\cite{li2018diffusion})
    \item EdgeGAT (Cora, Citeseer, Pubmed, 2 edge sensitive datasets Trade-B and Trade-M)
    \item HyperbolicGAT (Cora, Citeseer, Pubmed, Amazon Photo)
    \item HANs (ACM, IMDB, DBLP all requiring metapath information)
    \item SAN (CLUSTER, PATTERN)
    \item GraphBERT (Cora, Citeseer, PubMed)
    \item CAT (Cora, Cite, Pubmed, CoAuthorCS~\cite{pitfallsofgnnevaluation}, OGB-Arxiv\cite{hu2020open})
    
\end{itemize}

\subsection{Edge Level}
Edge level tasks involve prediction of the presence of edges between nodes or labelling existing edges, also commonly referred to as link classification~\cite{getoor2005link}. Knowledge graph problems often deal with tasks involving predicting missing relations or missing links between existing entities. 
\begin{itemize}
    \item GAT (OGB)
    \item GATv2 (OGB)
    \item HyperbolicGAT (realtional reasoning CLEVR, Sort-of-CLEVR)
    \item GAATs~\cite{wang2019knowledge}
    \item SAttLE~\cite{baghershahi2021self} (FB15k-237, WN18RR)
    \item HittER\cite{chen2021hitter} (FB15k-237, WN18RR, FreebaseQA, WebQuestionsQA)
\end{itemize}

\subsection{Graph Level}
These involve classification and regression tasks at the graph level eg. classifying if the graph of a certain molecule shows properties of inhibiting HIV or not.
\begin{itemize}
    \item SAN (ZINC regression, MolHIV, MolPCBA classification)
\end{itemize}

\subsection{Others}
These are two tasks that GraphBERT uses for pretraining: the node raw attribute reconstruction task focussing on extracting node attribute information and graph structure recovery task focuses on graph connection information.
\begin{itemize}
    \item SAN (ZINC regression, MolHIV, MolPCBA classification)
\end{itemize}

\begin{table*}[!ht]
    \centering
    \scalebox{0.75}{
    \begin{tabular}{l l l l l l l l l l l l l l}
    \hline
        Property & GAT & GT sparse & GT Full & SAN & Edge GAT & HAN &  Graph Transformer  & GraphBERT & GATv2 & CAT & Graphormers & Coarformer & LiteGT \\ \hline
        Preserves local structure in attention & {✓} & {✓} &   & {✓} & {✓} &   & {✓} & {✓} & {✓} & {✓} &   &   &   \\ 
        Uses edge features &   & {✓} &   & {✓} & {✓} &   & {✓} &   &   &   &   &   &   \\ 
        Connects non-neighbouring nodes &   &   & {✓} & {✓} &   &   &   & {✓} &   & {✓} &   &   &   \\ 
        Connect nodes in metapaths &   &   &   &   &   & {✓} &   &   &   &   &   &   &   \\ 
        Incorporate node type information &   &   &   &   &   & {✓} &   &   &   &   &   &   &   \\ 
        Uses PE for attention &   & {✓} & {✓} & {✓} &   &   & {✓} & {✓} &   &   &   &   &   \\ 
        Use a PE with structural information &   & {✓} &   & {✓} &   &   & {✓} & {✓} &   &   &   &   &   \\ 
        Aware of eigenvalue multiplicities &   &   &   & {✓} &   &   &   &   &   &   &   &   &   \\ 
        Invariant to the sign of the eigenvectors &   &   &   &   &   &   &   &   &   &   &   &   &   \\ 
        Consider higher-order neighbors &   &   &   &   &   &   &   & {✓} &   & {✓} &   &   &   \\ 
        Invariant to node ordering & {✓} &   &   &   & {✓} & {✓} &   &   & {✓} &   &   &  {✓} & {✓} \\ 
        Incorporate node centrality &  & &   &  &  &   &  &  &  &  &  {✓}  &   &    \\
        Incorporate spatial encoding  &  & &   &  &  &   &  &  &  &  & {✓}  &   &  \\
        Incorporate node sampling  &  & &   &  &  &   &  &  &  &  &   & {✓} & {✓} \\
    \end{tabular}}
\end{table*}

\section{Limitations of Message Passing Paradigms}
The dominant techniques in GNNs incorporate a sparse message-passing process to directly capture graph structure~\cite{hamilton2020graph} wherein messages are iteratively passed between nodes in the graph.~\cite{hamilton2020graph} provide a great review of different paradigms. However, this message-passing paradigm has been plagued with several limitations. eg. The expressiveness of message passing seems inescapably limited by the Weisfeiler-Lehman isomorphism hierarchy i.e. popular GNN models cannot distinguish between graphs indistinguishable by the 1-WL test~\cite{maron2019provably, morris2019weisfeiler,WLgoesneural,spectralnet}. 

Besides, message passing paradigms have been victims of \textit{oversquashing} and \textit{oversmoothing}:

With the exponential blow-up of computational routes, it becomes hard for graph neural networks to relay information to distant neighbors. This hardness is referred to as \textbf{\textit{oversquashing}}. 

With the addition of more number of layers, GNNs have not shown performance gains. This limitation is referred to as \textbf{\textit{oversmoothing}}~\cite{cai2020note}. 

Besides, the message passing paradigm limits the structure of the model's computation graph necessitating the need for approaches which provide the flexibility of soft-inductive bias. The Transformer architecture, for example, eliminates any structural inductive bias by encoding the structure with soft inductive biases like positional encodings~\cite{spectralnet}.

\section{Attention Architectures in Graph Neural Networks}
We first formulate and characterize the attention equations popular in sequential problems by~\cite{attentionIsAllYouNeed, bahdanau2014neural} and then describe graph variants of the same. 

In the transformers architecture, Vaswani et al,~\cite{attentionIsAllYouNeed} define the scaled dot-product attention for query, key and value matrices $Q$, $K$ and $V$ as follows:
\begin{equation*}
Attention(Q, K, V) = softmax(\frac{QK^T}{\sqrt{d_k}})V
\end{equation*}

Sequence-to-sequence models computed context representations $c_i$ for the $i$th decoder step by attending over all $T_x$ encoder steps indexed by $j$.
\begin{equation*}
c_i = \sum_{j=1}^{T_x} \alpha_{ij}h_j
\end{equation*}
where $\alpha_{ij}$ represented the learned attention weights
\begin{equation*}
\alpha_{ij} = \frac{\exp(e_{ij})}{\sum_{k=1}^{T_x} \exp(e_{ik})}
\end{equation*}
\begin{equation*}
e_{ij} = a(s_{i-1}, h_j)
\end{equation*}
A common choice for $a$ has been Bahdanau's attention~\cite{bahdanau2014neural} 
\begin{equation*}
e_{ij} = v^T tanh(W[s_{i-1}; h_j])
\end{equation*}

The above equations brought about unprecedented success for NLP tasks like machine translation, speech recognition, question answering, etc. and no wonder, the GNN community was motivated to incorporate the same to compute node representations. This was accomplished by ``attending'' over other nodes in the graph as we will see in subsequent sections.

\subsection{Graph Attention Networks (GAT)}
Veličković et al~\cite{velivckovic2017graph}'s seminal work established Graph Attention Networks, computing node representations by attending over neighbouring nodes $\mathcal{N}_i$ or nodes one-hop away. For every node $i$, each neighbouring node $j$ is weighted by a factor $\alpha_{ij}$ computed as:
\begin{equation*}
\vspace{0.1cm}
\alpha_{ij} =\frac{\exp(e_{ij})}{\sum_{k\in\mathcal{N}_i} \exp(e_{ik})}.
\end{equation*}
where $e_{ij}$ for nodes $i$ and $j$ can be expressed further in terms of their node features $h_i$ and $h_j$
\begin{equation}\label{alphaij}
	\alpha_{ij} = \frac{\exp\left(\text{LeakyReLU}\left(\vec{\bf a}^T[{\bf W}\vec{h}_i\|{\bf W}\vec{h}_j]\right)\right)}{\sum_{k\in\mathcal{N}_i} \exp\left(\text{LeakyReLU}\left(\vec{\bf a}^T[{\bf W}\vec{h}_i\|{\bf W}\vec{h}_k]\right)\right)}
\end{equation}
where $\|$ represents concatenation as described in their paper~\cite{velivckovic2017graph}.Figure ~\ref{figattn} describes the neighbourhood attention computation as described in~\cite{velivckovic2017graph}.

The final representation of the node was then computed by taking a linear weighted sum of the the neighbours as follows:
\begin{equation}\label{eqnatt}
	\vec{h}'_i = \sigma\left(\sum_{j\in\mathcal{N}_i} \alpha_{ij} {\bf W}\vec{h}_j\right).
\end{equation}
where $\sigma$ represents a non-linear activation function.

Taking inspiration from \cite{attentionIsAllYouNeed}, multi-head attention is similarly computed. This multi-head attention is found to stabilize learning~\cite{attentioningraphs}. Mathematically, Equation \ref{eqnatt} is executed $K$ times (the no. of attention heads), and concatenated (denoted by $\parallel$), resulting in the following output feature representation:
\begin{equation}\label{multiheadequation}
	\vec{h}'_i = \concat_{k=1}^K \sigma\left(\sum_{j\in\mathcal{N}_i}\alpha_{ij}^k{\bf W}^k\vec{h}_j\right)
\end{equation}
$k$ is used to denote the $k$-th attention head.

\subsection{Gated Attention Networks (GaAN)}
Zhang et al\cite{zhang2018gaan} hypothesize that some of the attention heads in~\ref{multiheadequation} might be redundant and could mislead final predictions. They strive to rectify this by introducing small convolutional subnetworks and compute a soft gate (0:low importance to 1:high importance) at each attention head to control the importance of that head:
\begin{equation}\label{gatedattention}
	\vec{h}'_i = {\bf W}(\vec{h}_i \oplus \concat_{k=1}^K \sigma \left(g_i^{(k)}\sum_{j\in\mathcal{N}_i}\alpha_{ij}^k{\bf W}^k\vec{h}_j\right))
\end{equation}
where $g_i^{(k)}$ is the value of the $k$th head at node $i$. 
\begin{equation}
{g}_i = [g_i^{(1)},..., g_i^{(K)}] = \psi_{g}(\mathbf{x}_i, \mathbf{z}_{\mathcal{N}_i})
\end{equation}
$\psi_{g}$ is a convolutional network that takes in the centre node and the neighbouring nodes as input and to compute gate values. The GaAN approach additionally adopted the key-value attention and the dot product attention as compared to GAT which does not compute separate value vectors.

\subsection{Edge GATs (EGATs)}
While the architectures discussed above only use node features in attention computation, some tasks might benefit with the incorporation of edge information. Eg. popular knowledge graphs like FreeBase and ConceptNet have lots of relations between  entities. Tasks like classifying entities (node classification) and predicting links between them would greatly benefit via cues of edge information. Trading networks too generally model send/receive payments as edges between nodes of users. Naturally, modelling relations has been an important facet of GNN research. Relational Graph Convolutional Networks (R-GCNs)~\cite{schlichtkrull2018modeling} introduced modelling relational data in GCNs for two
knowledge base completion tasks: link prediction \& entity classification. 

Chen \& Chen~\cite{chen2021edge} argue that different graphs may have different preferences for edges and weights and hence introduce Edge GATs. They extend the use of GATs to incorporate edge features in addition to the original node features.

Node representation described in ~\ref{alphaij} now additionally incorporates edge embeddings $e_{ik}$.\footnote{$W$ has been omitted to reduce space.}
\begin{equation}
   \alpha_{ij} = \frac{{\rm exp}({\rm LeakyReLU}(\vec{\textbf{a}}^T[\vec{h}_i \Vert \vec{h}_j \Vert \vec{e}_{ij}]))}{\sum_{k \in \mathcal{N}_i}{\rm exp}({\rm LeakyReLU}(\vec{\textbf{a}}^T[\vec{h}_i \Vert \vec{h}_k \Vert \vec{e}_{ik}]))}
\end{equation} 
Node updates are performed similar to GAT's Equation~\ref{eqnatt}. In a multi-layer attention architecture wherein multiple node encoders would be stacked, the last encoder attends over $\vec{h}_j \Vert \vec{e}_{ij}$ instead of $\vec{h}_j$.

Edge embeddings are then updated in parallel by attending over neighbouring edges. This is achieved by reversing the roles of nodes and edges and defining neighbouring edges as those connected with a common vertex. Effectively, node information is used to update edge representation. Arguably, such an update seems counter intuitive when graphs have independent pre-defined relations. It can be argued that certain edges might be highly prevalent in conjunction with specific subsets of nodes making node information vital for edge representation e.g. in graphs with a large number of relation types.

\begin{figure*}
    \centering
    \includegraphics[width=1\columnwidth, height=4cm]{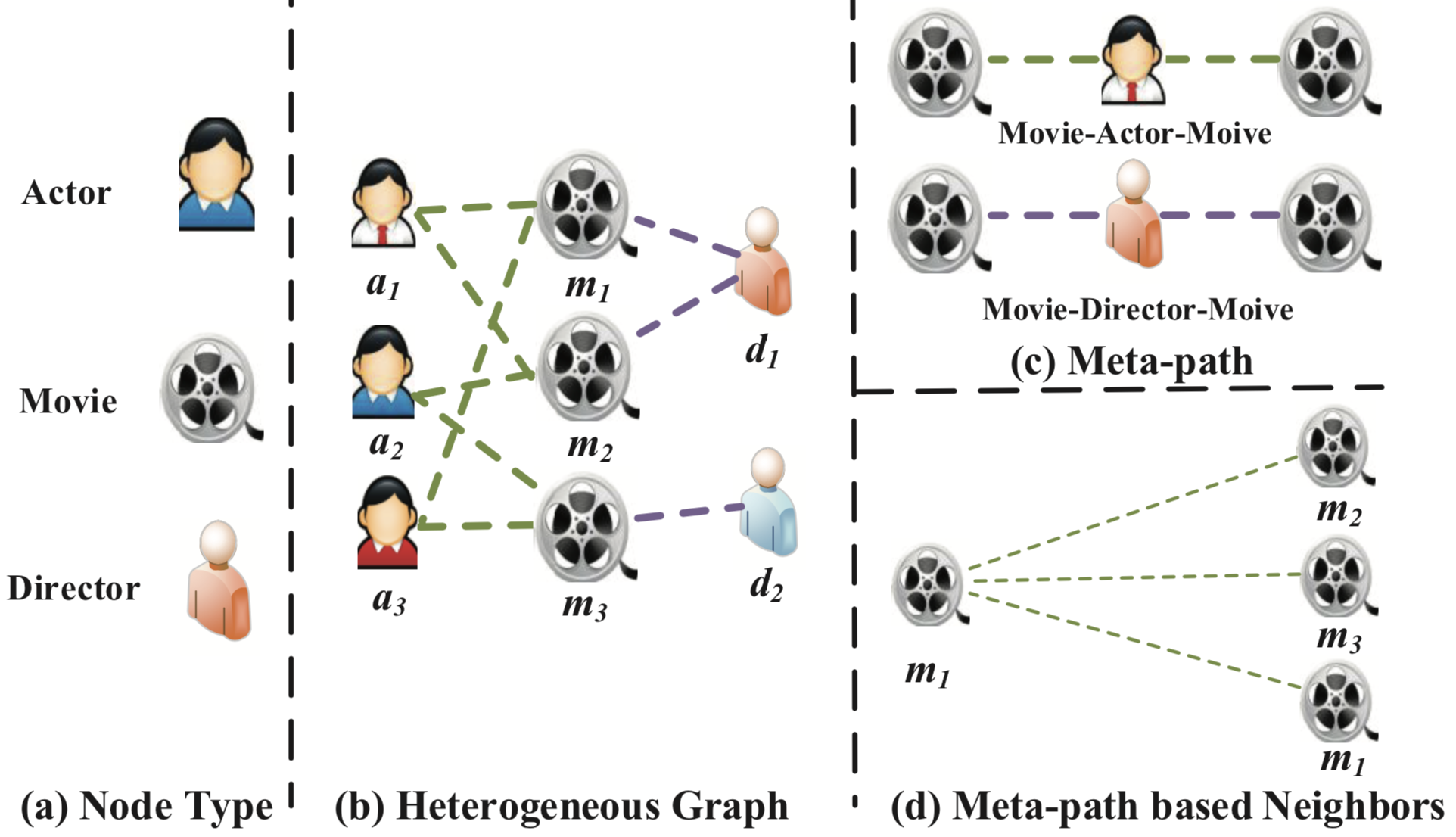}
    \caption{An example of a meta-path as  described in Wang et al~\cite{wang2019heterogeneous}}
    \label{fig:hetero}
\end{figure*}

\subsection{Heterogeneous Attention
Networks (HANs)}
Many real world graphs contain multiple types of nodes and edges as well as crucial information residing in the form of ``meta-paths''. Widely used in data mining, such heterogeneous graphs, adapted from heterogeneous information networks (HIN) contain more comprehensive information and richer semantics.~\cite{wang2019heterogeneous, shi2016survey}. Eg. one key feature of HINs is the ability to spread information through various edges among different-typed nodes~\cite{shi2016survey}  i.e depending on the meta-paths, the relation between nodes
in a heterogeneous graph can have varying semantics~\cite{wang2019heterogeneous}. Figure~\ref{fig:hetero} describes how the relation between two movies can be described by two metapaths: Movie-Actor-Movie or Movie-Director-Movie.

~\cite{wang2019heterogeneous} introduce hierarchical attention which includes 1) node level attention attending over the meta-path based neighbours and 2) semantic-level attention to learn the importance of different metapaths.

\textbf{Type-Specific Transformation}: The authors of HAN first project the features of different types of nodes into the same
feature space by designing a node type-specific transformation matrix:

\begin{equation}
\mathbf{h}_i'= \mathbf{M}_{\phi_i} \cdot \mathbf{h}_i
\end{equation}

\textbf{Attention within metapath neighbours}: For a given metapath $\Phi$, node level attention  is computed is computed over all the nodes ${N}_i^{\Phi}$ i.e. nodes appearing in that metapath. This attempts to record the influence of each node in the metapath.

\begin{equation*}\label{metapathattention}
\mathbf{z}^{\Phi}_i=
\overset{K}{\underset{k=1}{\Vert}}
\sigma \biggl( \sum_{j \in \mathcal{N}_i^{\Phi}} \alpha_{ij}^{\Phi} \cdot \mathbf{h}_j'  \biggr).
\end{equation*}

where $\mathcal{N}_i^{\Phi}$ represents all nodes appearing on the metapath $\Phi$. Attention weights $\alpha_{ij}^{\Phi}$ are computed just like the previously described softmax ( Eq.~\ref{alphaij}) whilst attending over metapath neighbours rather than immediate neighbours.

\textbf{Attention over metapath-specific node representations}: Since each metapath from $\left\lbrace \Phi_1,\ldots,\Phi_{P}\right\rbrace $ presents its own semantics, the node representation is computed for each metapath providing $P$ groups of node representations denoted as $\left\lbrace \mathbf{Z}_{\Phi_1},\ldots,\mathbf{Z}_{\Phi_{P}}\right\rbrace $. Semantic-level attention~\cite{wang2019heterogeneous} refers to attending over the node representations of these metapaths. 
\begin{equation}
\mathbf{Z}=\sum_{p=1}^{P} \beta_{\Phi_p}\cdot \mathbf{Z}_{\Phi_p}.
\label{sem_agg}
\end{equation}
where the semantic attention weights $\beta_{\Phi_p}$ are computed as follows 

\begin{equation*}
w_{\Phi_p} =\frac{1}{|\mathcal{V}|}\sum_{i \in \mathcal{V}} \mathbf{q}^\mathrm{T} \cdot \tanh(\mathbf{W}\cdot \mathbf{z}_{i}^{\Phi_p}+\mathbf{b})
\end{equation*}
\begin{equation*}
\beta_{\Phi_p}=\frac{\exp(w_{\Phi_p})}{\sum_{p=1}^{P} \exp(w_{\Phi_p})} ,
\end{equation*}

\subsection{Hyperbolic GATs}
The above networks that we looked at were designed primarily for the Euclidean space. However, some work has pointed out that Euclidean spaces may not provide the perfect geometric environment for learning graph representations as graphs exhibit many non-Euclidean traits\cite{wilson2014spherical, bronsteinhyperbolic, krioukov2010hyperbolic,liu2019hyperbolic}. Some graphical properties like hierarchical \& power-law structure are naturally reflected in hyperbolic spaces~\cite{krioukov2010hyperbolic,liu2019hyperbolic} and hence a lot of subsequent work focused on graph neural networks in the hyperbolic space~\cite{zhanghyperbolicgraphattention, gulcehre2018hyperbolic}.

Hyperbolic Graph Attention Network (Zhang et. al~\cite{gulcehre2018hyperbolic}): Since basic algebraic operations like addition \& multiplication are not straightforward in the hyperbolic space,~\cite{zhanghyperbolicgraphattention} introduce a hyperbolic proximity based attention mechanism, Hyperbolic Attention Network (HAN) by utilizing gyrovector spaces to featurise the graph. Gyrovector spaces, introduced by~\cite{ungar2008gyrovector} provide an elegant formalism for algebraic operations in hyperbolic geometry. The authors notice that HAT performs better in most problems, especially in low dimensional settings.

While there are plenty of GNN architectures proposed in the hyperbolic space, it would be infeasible to discuss and introduce a separate parallel set of notation of hyperbolic geometry within the scope of this review. We hence encourage the reader to read up the article by ~\cite{bronsteinhyperbolic}.

\subsection{Graph Transformers}
Along with attention, positional encodings have been extensively studied for sequential problems where positional information of words is crucial eg. tasks in NLP. Analogously, for fundamental graph tasks, recent studies~\cite{dwivedi2020benchmarkgnns,dwivedi2020benchmarkgnns,you2019position,ying2021transformers} point out positional information to be key to improve and overcome many of GNNs' failures .~\cite{dwivedi2020generalization} replace the sinusoidal embeddings commonly used for sentences i.e. line graphs with their generalised counterparts  - Laplacian positional encodings. They also substitute layer normalisation with batch normalisation. Additionally, the architecture attends over neighbouring nodes and takes edge representation too into account.

\subsection{Graph BERT}
Li et al. attempt to incorporate global information of the graph by extending the attention computation over all the nodes i.e. not just neighbouring nodes. However, such an approach can be costly and does not take advantage of graph sparsity. Besides, GNNs have been known to suffer from problems of ``suspended animation'' \& ``oversmoothing''. To rectify such issues, [GraphBERT] performs sampling of linkless subgraphs i.e. sample nodes together with their context.
While incorporating global information of the graph, it can be crucial to understand the position of nodes within the broader context of the graph. \cite{you2019position} argue that two nodes residing in very different parts of the graph
may have an isomorphic topology of neighbourhood but may deserve different representations. They incorporate every node’s positional
information by computing the distance from a set of random nodes in each forward pass. 

Importantly, domain specific \& low-resource graphs might not be large enough to train such parameter-heavy graph neural networks and hence learning strategies, successful in machine translation like pre-training + fine-tuning [BERT] can prove beneficial. GraphBERT has been introduced as a graph representation learning on the same lines. The authors pretrained GraphBERT for two tasks, namely node attribute reconstruction and graph structure recovery \& fine-tuned it for the task of node classification.

A D-layer GraphBERT architecture is summarised below:
\begin{equation*}
\begin{cases}
\vspace{5pt}
\mb{H}^{(0)} & \hspace{-10pt} = [\mb{h}_i^{(0)}, \mb{h}_{i,1}^{(0)}, \cdots, \mb{h}_{i,k}^{(0)}]^\top ,\\
\vspace{5pt}
\mb{H}^{(l)} &\hspace{-10pt}= \mbox{G-Transformer} \left( \mb{H}^{(l-1)}\right), \forall l \in \{1, 2, \cdots, D\},\\
\mb{z}_i &\hspace{-10pt}= \mbox{Fusion} \left( \mb{H}^{(D)} \right). 
\end{cases}
\end{equation*}
where each $\mb{h}_i^{(0)}$ is a concatenation of 4 different types of embeddings for a node: 1) the raw feature
vector embeddings, (2) Weisfeiler-Lehman absolute role embeddings, (3) intimacy based relative positional embeddings, \& (4) hop based relative distance embeddings. And $G-Transformer$ is the usual softmax attention alongwith a residual term $\mbox{G-Res}$ as defined in graph residul networks~\cite{zhang2019gresnet}.
\begin{equation*}
\begin{aligned}
\hspace{-5pt} \mb{H}^{(l)} &= \mbox{G-Transformer} \left( \mb{H}^{(l-1)}\right)\\
&= \mbox{softmax} \left(\frac{\mb{{H}^{(l-1)}{W}_Q^{(l)}} \mb{{H}^{(l-1)}{W}_K^{(l)}}^\top}{\sqrt{d_h}} \right) \mb{{H}^{(l-1)}{W}_V^{(l)}} \\& +           \mbox{G-Res} \left( \mb{H}^{(l-1)}, \mb{X}_i\right),
\end{aligned}\vspace{-2pt}
\end{equation*}

Laplacian positional encodings outperform the WL-positional encodings introduced above in capturing positional and structural information as well as in generalization.

\subsection{Spectral Attention Network (SAN)}
Taking inspiration from spectral graph theory, SAN~\cite{spectralnet} try to address some of the theoretical limitations of the above Graph Transformer work. They make use of the complete Laplace spectrum for positional encodings. Apart from the benefits offered by~\cite{dwivedi2020generalization}, SAN incorporate variable number of eigenvectors, the whole spectrum of eigenvalues and are aware of eigenvalue multiplicities.
Given appropriate parameters and the utilization of the whole set of eigenfunctions, SAN can approximately differentiate any pair of non-isomorphic graphs, making it more powerful than any WL test. SAN also might be less prone to oversquashing.

The authors argue that unlike the graph transformers discussed earlier which don't exploit eigenvalues and eigenfunctions wholly, SAN is able to model physical interactions better i.e. interactions commonly observed in physics, biology and images. SAN is seen to outperform other attention-based architectures by a large margin on the tasks of ZINC~\cite{jin2018junction}, a molecular graph regression dataset, PATTERN \& CLUSTER~\cite{dwivedi2020benchmarkgnns}, two synthetic benchmarks for node classification \& MolPCBA~\cite{hu2020open}, a dataset for molecular graph classification.

\subsection{Gated Attention Network v2}
~\cite{brody2021attentive} argued that the previous well-known GAT design proposed by Veličković et al,\cite{velivckovic2017graph} and its variants which spread out across different graph domains computed a limited form of attention which was static in nature rather than the actual expressive attention function of Bahadanu et al~\cite{bahdanau2014neural}. ~\cite{brody2021attentive} showed that the attention function is monotonic with respect to the neighbouring nodes. This monotonicty is shared across all nodes in the graph without being conditioned on the query node. GATv2 instead computes dynmamic attention, with a simple fix by switching
the order of internal operations in GAT.

\begin{align*}
&			 \mathrm{GAT} \text{ \cite{velivckovic2017graph}:} &
e\left(\vec{h}_i, \vec{h}_j\right)= &
	\text{LeakyReLU}\left(\vec{\bf a}^T[{\bf W}\vec{h}_i\|{\bf W}\vec{h}_j]\right)
\\
	&			 \mathrm{GATv2} \text{\cite{brody2021attentive}:} &
e\left(\vec{h}_i, \vec{h}_j\right) = &
\vec{\bf a}^T
	\text{LeakyReLU}\left([{\bf W}\vec{h}_i\|{\bf W}\vec{h}_j]\right)
\end{align*}

The standard GAT scoring function\cite{velivckovic2017graph} applies the learned layers {\bf W} and $\vec{\bf a}^T$ consecutively, and effectively collapses them into one linear layer.
By simply applying the $\vec{\bf a}^T$ layer after the non-linearity ($\mathrm{LeakyReLU}$) and the {\bf W} layer after the concatenation, GATv2 overcomes this issue rendering it with a universal approximator function and making it strictly more expressive than GAT. Besides theoretical superiority, GATv2 has shown empirical advantages over GAT on various tasks which require dynamic selection of nodes.

\subsection{Graph Conjoint Attention
networks (CATs)}
Contextual interventions have been found as helpful external elements that may increase attention and cognitive capacities in cognitive science~\cite{jones2004joint}. Inspired by this finding, \cite{he2021learning} describe the concept of conjoint attentions. They incorporate node cluster embedding, and higher-order structural correlations, arguing that such external components can enhance learning and provide more robustness to graph neural networks, e.g. against overfitting.

Here is the formal definition of a structural intervention: $\Psi(\cdot)$ can be any distance function (eg. Euclidean) and $\phi(\cdot)$ can be any operator transforming $\mathbf {C}$ to the same dimensionality of $\mathbf Y$, the prior feature matrix. The structural intervention between two nodes can be defined as below:
\begin{equation*}\label{gen}
	\begin{aligned}
		\mathbf {C}_{ij} = \mathop{\arg \min}_{\phi(\mathbf {C})_{ij}} \Psi(\phi(\mathbf {C})_{ij},\mathbf Y_{ij}),
	\end{aligned}
\end{equation*}
\begin{equation*}
	s_{ij} = \frac{\exp{(\mathbf C_{ij})}}{\sum_{k \in \mathcal N_i} \exp{(\mathbf C_{ik})}}.
\end{equation*}

Given the $f_{ij}$ (Equation~\ref{alphaij}) and $s_{ij}$, \cite{he2021learning} propose two different strategies to compute the Conjoint Attention scores, aiming at allowing CATs to depend on the structural intervention at different levels.
The first mechanism is referred here as \textit {Implicit direction}.
Each CA layer introduces two learnable parameters, $g_f$ and $g_s$. They can be obtained in the following fashion and are used to determine the relative importance of feature and structural correlations:
\begin{equation*}\label{pen}
	r_f = \frac{\exp{(g_f)}}{\exp{(g_s)+\exp{(g_f)}}}, r_s = \frac{\exp{(g_s)}}{\exp{(g_s)+\exp{(g_f)}}},
\end{equation*}
CAT subsequently computes the attention score based as follows:
\begin{equation*}\label{att-ip}
	\alpha_{ij} = \frac{r_f\cdot f_{ij}+r_s\cdot s_{ij}}{\sum_{k \in \mathcal N_i}[r_f\cdot f_{ik}+r_s\cdot s_{ik}]}=r_f\cdot f_{ij}+r_s\cdot s_{ij}.
\end{equation*}
The explicit strategy performs a different computation as follows:
\begin{equation*}\label{att-dp}
	\alpha_{ij} = \frac{f_{ij}\cdot s_{ij} }{\sum_{k \in \mathcal N_i}f_{ik}\cdot s_{ik}}.
\end{equation*}
Eventually, the higher layers are updated as follows, with a learnable parameter $\epsilon \in (0, 1)$ added to introduce expressivity of the CAT approach:
\begin{equation}\label{att-aggregation}
	\mathbf h^{l+1}_i = (\alpha_{ii}+\epsilon\cdot\frac{1}{\vert \mathcal N_i \vert})\mathbf {W}^l\mathbf h^l_i+ \sum_{j \in \mathcal N_i, j \ne i} \alpha_{ij} \mathbf {W}^l\mathbf h^l_j,
\end{equation}
The authors showed CAT to be better than GAT on many node classification and clustering tasks but the approach seems to increase both space and time complexity.

\subsection{Additional Attention Based GNNs}
Additional variants majorly build on top of these fundamental equations. Eg, Graphormers~\cite{ying2021transformers} encode node centrality and spatial relations encoded in node pairs and edges alongwith the softmax attention. To improve efficiency, Coarformer~\cite{anonymous2022coarformer} performs attention on a courser version of the original graph.~\cite{sublineargraphcoursening} give an account of sublinear graph coarsening strategies to reduce the number of nodes by up to a factor of ten without causing a noticeable performance degradation. LiteGT~\cite{litegt} describes $O(nlogn)$ time node sampling strategies resulting in a 100x time reduction and reduced model size to enjoy similar performance.


\section{Summary}
Now that we've witnessed the mathematical details of each architecture from the point of view of the attention function and downstream application, we can revisit these architectures to address the task at hand. For node classification problems where node positions might not play a big role, it would be worthwhile experimenting with Graph Attention Networks, GAT and GATv2 by attending over neighbouring nodes. 
While many graphs are sparse in nature while also require making global inferences, it might be necessary to look at the whole graph but might be too costly to do so. In such a case, it is imperative to investigate some of the sampling techniques used to prune the attention candidates as well as encode positional information. eg. as used in GraphBERT alongwith retrieving positional encodings. Having a more exhaustive set of positional encodings and performing better than GraphTransformer, SAN could also be a vital choice in such problems.

For tasks like link prediction or other graph tasks reliant on graphs with vital edge information, it is worthwhile incorporating edge information like EdgeGATs. Besides, one has to be cautious with these higher attention variants as each of these architectures when employed with multiple attention heads would charge a heavy computational cost. Eg, EdgeGATs and CATs, despite showing impressive performance might not be a good choice when computational budgets are constrained. Graphormers would be a great choice for tasks with smaller knowledge graphs since the complexity grows quadratically.

As mentioned earlier, architectures may perform well depending on the graph at hand. There is hardly any universal architecture suitable for every problem. We hope this tutorial makes readers better informed towards their design choices of GNN architectures.

\section{Other Avenues in Graph ML}
Despite attention variants being the focus of the current study, there are other sub-fields of graph machine learning eyeing different areas of GNNs which deserve equal attention. 
Other innovative directions in the field of graph machine learning can be attributed to the works of GFlowNets~\cite{bengio2021gflownet, bengio2021flow}, the study of how GNNs are aligned with dynamic programming~\cite{Xu2020What}, GNNs with combinatorial optimisation~\cite{cappart2021combinatorial}, Satorras, et al~\cite{satorras2021n}'s Equivariant GNNs, Klicpera et al~\cite{klicpera2021gemnet}'s GEMnet, etc.

\section{Conclusion}
This tutorial attempts to delve into multiple attention architectures detailing the high-level mathematical details. We provide the crux of each of the attention equations in a uniform notation for the benefit of readers. We hope it provides guidance to researchers dealing with graph-structured problems so that they can get a high-level overview for their tasks.

\section{Acknowledgements}
We would like to thank the anonymous reviewers from the CS-570 class of Emory University and Prof. Jinho D. Choi for their useful suggestions. The figures of the animated characters are sourced from the webpages of (inline linked)  \href{https://easydrawingtutorials.com/index.php/misc/189-draw-jerry}{Jerry}, \href{ https://www.pinterest.co.uk/pin/346143921342318543/}{Tom}, \href{https://www.deviantart.com/gamerfaceswaper/art/Tom-and-Jerry-Face-Swap-882945324}{mixed mouse} and \href{https://www.freesvgdownload.com/minnie-mouse-head-svg-png-free-download/}{Minnie}.

\bibliography{bibliography}
\bibliographystyle{plainnat}

\end{document}